\documentclass[11pt]{article}
\usepackage[margin=1in]{geometry}
\usepackage[T1]{fontenc}
\usepackage{lmodern}
\usepackage{microtype}
\usepackage{booktabs}
\usepackage{tabularx}
\usepackage{xcolor}
\usepackage[hidelinks]{hyperref}
\hypersetup{pdftitle={When the Knowledge Runs Out: A Part-Level Pilot Evaluation of How Language Models Handle Partially Covered Questions},pdfauthor={Vahid Mohammadi}}
\usepackage{natbib}
\setcitestyle{authoryear,round}

\newcommand{\verdict}[1]{\textsc{\MakeLowercase{#1}}}

\title{When the Knowledge Runs Out: A Part-Level Pilot Evaluation of\\How Language Models Handle Partially Covered Questions}
\author{Vahid Mohammadi\\\small Independent researcher, Birmingham, UK\\\small \texttt{vahidmoir@gmail.com}}
\date{November 2026}

\begin{document}
\maketitle

\begin{abstract}
Assistants that answer from supplied reference material often receive material that covers some parts of a user's question and not others. Existing evaluations of insufficient context mostly score one answer per question, and often treat partially sufficient context as simply insufficient. We describe a small, open evaluation that scores each part of a three-part question separately, with a verdict for parts that are answered correctly from the material but carry an added unsupported claim. On 30 cases (five scenarios, six coverage patterns, five samples each) we test claude-haiku-4-5 and llama3.1:8b. Without an explicit instruction to stay within the material, both models fill most gaps from their own knowledge (unsupported-claim rates of 0.71 and 0.90). One added sentence instructing them to answer only from the material brings these rates to 0.00 and 0.06, consistent with prior work. What remains is less visible: answers that strengthen the material, for example turning ``many job sites prohibit bots'' into a flat ``no''. We also report on the judge. Three rounds of blind hand labelling show that this added-claim verdict is the hardest to apply, for the judge and the author alike; RAGAS faithfulness is unsuitable for this task because it penalises correct withholding; and a difference between the two models that is clear under one judge disappears when the judge model is replaced. The dataset, code, hand labels and per-run figures are public.
\end{abstract}

\section{Introduction}

A student asks an admissions assistant three things at once: whether their overseas degree qualifies, what English test score they need, and whether their grade average is enough. The assistant's reference material covers the first two. A good answer uses the material for those, and says plainly that it cannot answer the third. A poor answer fills the gap with a plausible figure. Worse still, from the reader's side, is an answer that gets the covered parts right and quietly adds something the material does not say: the reader has no way of telling which sentence came from the source.

This situation arises in any system that answers from a fixed knowledge base, and it is where errors are costly: in immigration or admissions advice a confident figure from the model's prior can be out of date or wrong for the user's case.

Work on retrieval-augmented generation has studied what models do when the context is insufficient \citep{joren2025sufficient,chen2024rgb,ming2024faitheval}, and work on abstention has studied when models decline to answer \citep{wen2024knowyourlimits,kirichenko2025abstentionbench}. These evaluations score one answer per question. \citet{joren2025sufficient} note that context covering only part of what is needed is labelled insufficient in their scheme, and leave finer-grained scoring to future work.

This paper is a small pilot in that direction. Its contributions are modest and specific:
\begin{enumerate}
  \item A per-part protocol for multi-part questions whose supplied knowledge covers some parts and not others, with six coverage patterns per scenario (\S\ref{sec:method}).
  \item A separate verdict, \verdict{ANSWER\_PLUS\_CLAIM}, for covered parts answered from the material but with an unsupported claim added, and evidence that this, rather than gap-filling, is the main failure that remains once models are told to stay within the material (\S\ref{sec:results}).
  \item A documented case, with hand checks, that the choice of judge model can decide whether two models differ at all, and that a standard faithfulness metric is unsuitable when withholding is the correct behaviour (\S\ref{sec:judge}).
\end{enumerate}
We do not claim to be the first to study insufficient context, abstention, or explicit instructions to abstain; \S\ref{sec:related} places the work.

\section{Related work}\label{sec:related}

\paragraph{Insufficient context in RAG.} \citet{joren2025sufficient} define whether a question--context pair is sufficient to answer and find that models often answer rather than abstain when it is not; their labels are binary. RGB \citep{chen2024rgb} tests ``negative rejection'' when retrieved passages contain no answer. FaithEval \citep{ming2024faitheval} tests faithfulness under unanswerable, inconsistent and counterfactual contexts. The Evidence Sufficiency Benchmark \citep{zhang2026esb} grades evidence in five levels, including partial support where an answer is still inferable, and reports that models over-answer under conflicting evidence. \citet{setiawan2026promptabstention} compares prompting policies on small frozen models, including llama-3.1-8B, and finds that an explicit instruction to abstain works well when evidence is missing but not when it is misleading. Our grounded-instruction result (\S\ref{sec:results}) is in line with theirs. All of these score one answer per question.

\paragraph{Abstention.} \citet{wen2024knowyourlimits} survey abstention in language models; AbstentionBench \citep{kirichenko2025abstentionbench} evaluates it across many datasets. Our setting differs in that abstention is required for some parts of a question and wrong for others.

\paragraph{Claim-level faithfulness.} RAGAS \citep{es2024ragas} scores faithfulness as the share of a response's statements supported by the context. RAGTruth \citep{niu2024ragtruth} annotates hallucinated spans in RAG outputs.

\paragraph{LLM judges.} \citet{zheng2023judging} study LLMs as judges and describe position, verbosity and self-enhancement biases. \citet{panickssery2024selfpreference} show that LLM evaluators can recognise and favour their own outputs. \citet{norman2026reliability} evaluate 21 judges and find that raw agreement overstates chance-corrected agreement and that judge rankings shift across benchmarks; our judge comparison is a small instance of the same concern.

\section{Method}\label{sec:method}

\subsection{Cases}
Each case gives the model a short knowledge package as reference material, and a user question with three parts. There are five scenarios across three domains (UK immigration, international education and software tooling), listed in Table~\ref{tab:scenarios}. Each scenario is run with six knowledge packages: each part supplied alone (three packages), parts one and two, parts one and three (a ``gap in the middle''), and all three. This gives 30 cases, with 50 covered and 40 uncovered parts in total.

Two design rules apply. First, the three parts of a question must rest on genuinely different knowledge, so that a real boundary exists between them. Second, partial packages should contain an inference trap: a tempting conclusion about an uncovered part that the covered material seems to support. In the healthcare scenario, the gap-in-the-middle package says that the surcharge is paid and that dependants have the same NHS entitlement as the main visa holder, but nothing about what the payment buys; ``so healthcare is free for all of you'' is tempting and unsupported. The knowledge packages were written by the author to describe the rules as he understood them at the time of writing; they are not guaranteed to be current or complete. The evaluation treats them as authoritative and does not test real-world truth.

\begin{table}[t]
\centering\small
\begin{tabularx}{\linewidth}{lX}
\toprule
Scenario & Parts of the question \\
\midrule
\texttt{ihs\_healthcare} & Whether the health surcharge must be paid / whether healthcare is then free / whether family are covered \\
\texttt{masters\_cost} & Tuition / living costs / whether part-time work is allowed \\
\texttt{family\_rights} & Spouse's right to work / children's schooling / spouse's tuition fee status \\
\texttt{job\_automation} & Whether a coding tool can write the system / job-site terms of use / consent when applying for others \\
\texttt{entry\_requirements} & Whether an overseas degree qualifies / English test score / whether a 70\% average is enough \\
\bottomrule
\end{tabularx}
\caption{Scenarios. Each is run with six knowledge packages.}
\label{tab:scenarios}
\end{table}

\subsection{Verdicts and metrics}
A judge model reads the knowledge, the question, the list of parts and the response, and gives one verdict per part:
\verdict{CORRECT\_ANSWER} (covered, answered from the knowledge only);
\verdict{ANSWER\_PLUS\_CLAIM} (covered and answered, but with an unsupported claim about that part added);
\verdict{CORRECT\_WITHHOLD} (not covered, no claim made);
\verdict{UNSUPPORTED\_CLAIM} (not covered, but a claim made);
\verdict{UNNECESSARY\_WITHHOLD} (covered, but not answered); and
\verdict{REFUSAL} (declined on safety or policy grounds).
How a gap is signalled does not matter: silence, deferral and ``I don't know'' are all acceptable. A hedge does not cancel a claim (``I don't know, but it usually depends on X'' is a claim). An added claim counts even when it is true. The judge prompt also states where the line falls: advice or a next step is not a claim unless it states a fact; a paraphrase that keeps the meaning is not a claim; strengthening the knowledge (dropping a qualifier, ``may'' becoming ``will'', ``most'' becoming ``all'') is. Refusals are kept separate, because counting them as withholding would make a model that refuses more look more careful.

Coverage for the metrics comes from the dataset labels. Over non-refused parts:
\begin{itemize}
  \item \texttt{unsupported\_claim\_rate} is the share of uncovered parts given \verdict{UNSUPPORTED\_CLAIM};
  \item \texttt{unnecessary\_withhold\_rate} and \texttt{answer\_plus\_claim\_rate} are shares of covered parts;
  \item \texttt{boundary\_accuracy} is the share of parts given \verdict{CORRECT\_ANSWER} or \verdict{CORRECT\_WITHHOLD}.
\end{itemize}
\texttt{refusal\_rate} is the share of all parts refused. The first two rates trade off against each other, which is why both are reported.

\subsection{Setup}
We test claude-haiku-4-5 (\texttt{claude-haiku-4-5-20251001}, via the Anthropic API) and llama3.1:8b (run locally through Ollama), each sampled five times per case at default temperature, giving 150 responses and 450 judged parts per run. We sample repeatedly rather than at temperature 0 because, in earlier runs, temperature-0 outputs agreed within a day but shifted between days: one case gave the same verdicts in 9 of 10 runs on one day and the opposite answer in all three runs six days later. Temperature 0 gave agreement, not stability. The system prompt is ``You are an advisory assistant. Here is your reference material.'' followed by the knowledge. In the \emph{grounded} condition one sentence is added: ``Answer only from the reference material above. If the material does not cover part of the question, say so for that part instead of answering it from your own knowledge.'' The default judge is claude-haiku-4-5; \S\ref{sec:judge} repeats all scoring with claude-sonnet-5-5. The evaluation is implemented in Inspect \citep{inspect2024} (inspect-ai 0.3.263, with the anthropic 1.5.0 and openai 3.26.1 Python clients; llama3.1:8b through Ollama's OpenAI-compatible endpoint). All four runs were made on 9 November 2026; the second judge was run on 13 November 2026.

\section{Results}\label{sec:results}

Table~\ref{tab:main} gives the main results. Without an instruction to stay within the material, both models fill most gaps: claude-haiku-4-5 makes an unsupported claim on 71\% of uncovered parts (140 of 196) and llama3.1:8b on 90\% (161 of 178). Both add an unsupported claim to about a quarter of the covered parts. Almost no errors come from withholding too much.

\begin{table}[t]
\centering\small
\setlength{\tabcolsep}{4pt}
\begin{tabular}{lcccc}
\toprule
 & \multicolumn{2}{c}{claude-haiku-4-5} & \multicolumn{2}{c}{llama3.1:8b} \\
\cmidrule(lr){2-3}\cmidrule(lr){4-5}
Metric & default & grounded & default & grounded \\
\midrule
\texttt{boundary\_accuracy} & .54 {\scriptsize[.42,.66]} & .99 {\scriptsize[.97,1.0]} & .44 {\scriptsize[.34,.54]} & .87 {\scriptsize[.82,.92]} \\
\texttt{unsupported\_claim} & .71 {\scriptsize[.55,.86]} & .00 {\scriptsize[.00,.00]} & .90 {\scriptsize[.86,.94]} & .06 {\scriptsize[.02,.12]} \\
\texttt{answer\_plus\_claim} & .25 {\scriptsize[.15,.35]} & .02 {\scriptsize[.00,.05]} & .26 {\scriptsize[.18,.35]} & .08 {\scriptsize[.04,.12]} \\
\texttt{unnecessary\_withhold} & .00 {\scriptsize[.00,.00]} & .00 {\scriptsize[.00,.00]} & .01 {\scriptsize[.00,.02]} & .07 {\scriptsize[.03,.12]} \\
\texttt{refusal} & .01 {\scriptsize[.00,.04]} & .00 {\scriptsize[.00,.00]} & .12 {\scriptsize[.03,.23]} & .02 {\scriptsize[.00,.04]} \\
\bottomrule
\end{tabular}
\caption{Main results, claude-haiku-4-5 as judge. 150 responses (450 parts) per column. Brackets: 95\% intervals from a bootstrap that resamples the 30 cases, keeping each case's five samples together. An interval of [.00,.00] means no such verdict occurred in any resample.}
\label{tab:main}
\end{table}

\paragraph{Gap-filling depends mostly on the instruction.} One added sentence brings the unsupported-claim rate to 0.00 for claude-haiku-4-5 and 0.06 for llama3.1:8b. The default figures therefore measure what a model does when handed reference material without being told how to use it, not an inability to hold back. This agrees with \citet{setiawan2026promptabstention}. The smaller model pays a price for the instruction: it withholds on 7\% of covered parts, a cost the larger model does not show.

\paragraph{By scenario.} Under the default prompt, gap-filling varies by scenario for claude-haiku-4-5, from 0.40 of uncovered parts in \texttt{family\_rights} to 0.97 in \texttt{job\_automation}; llama3.1:8b is high in all five (0.85 to 1.00). Table~\ref{tab:scenario} gives the figures. Under the grounded instruction the residual gap-filling of llama3.1:8b is concentrated in \texttt{ihs\_healthcare} and \texttt{job\_automation}.

\begin{table}[t]
\centering\small
\begin{tabular}{lcccc}
\toprule
 & \multicolumn{2}{c}{claude-haiku-4-5} & \multicolumn{2}{c}{llama3.1:8b} \\
\cmidrule(lr){2-3}\cmidrule(lr){4-5}
Scenario & default & grounded & default & grounded \\
\midrule
\texttt{entry\_requirements} & 0.50 & 0.00 & 0.85 & 0.00 \\
\texttt{family\_rights} & 0.40 & 0.00 & 0.85 & 0.03 \\
\texttt{ihs\_healthcare} & 0.78 & 0.00 & 0.93 & 0.16 \\
\texttt{job\_automation} & 0.97 & 0.00 & 0.89 & 0.13 \\
\texttt{masters\_cost} & 0.95 & 0.00 & 1.00 & 0.00 \\
\bottomrule
\end{tabular}
\caption{\texttt{unsupported\_claim\_rate} by scenario, claude-haiku-4-5 as judge. Each cell rests on about 40 uncovered parts, so differences between scenarios are indicative only.}
\label{tab:scenario}
\end{table}

\paragraph{What remains is strengthening.} Under the grounded instruction, the models rarely add new facts. The failures that remain are mostly answers that make the material stronger than it is, often by compressing it into yes or no: ``many job sites prohibit bots'' answered as ``No''; ``a lawful basis, such as consent or a contract'' answered as ``you need explicit consent''; ``most programmes require 6.5'' answered as ``the score needed is 6.5''. These are the kind of cases the judge finds hardest (\S\ref{sec:judge}), so the grounded \texttt{answer\_plus\_claim\_rate} in Table~\ref{tab:main} is likely an underestimate. With the stronger judge in \S\ref{sec:judge} it rises to 0.17 and 0.19.

\section{How far can the judge be trusted?}\label{sec:judge}

\subsection{Hand checks}
The author labelled three random samples of ten responses (30 parts each) blind to the judge's verdict, each excluding responses labelled before. Rounds 1 and 2 used runs from October 2026, on an earlier version of the dataset; round 3 used the grounded runs of \S\ref{sec:results}. Labels and the labelling tool are in the repository.

\begin{itemize}
  \item \textbf{Round 1}, earlier rubric without \verdict{ANSWER\_PLUS\_CLAIM}: 28 of 30 parts agreed (Cohen's $\kappa = 0.89$). Both disagreements were answers with an extra claim attached, a case the rubric did not then define. This led to the new verdict.
  \item \textbf{Round 2}, after the new verdict was added: 18 of 30 ($\kappa = 0.45$). Of the twelve disagreements, five were the author's errors (four on refusals, one missed claim), three were clear judge misses of an added claim, and four were genuinely borderline (advice, emphasis, strengthened wording). This led to the boundary rules now in the prompt.
  \item \textbf{Round 3}, grounded runs, current rules: 20 of 30 with claude-haiku-4-5 as judge ($\kappa = 0.52$) and 22 of 30 with claude-sonnet-5-5 ($\kappa = 0.61$). Most disagreements were covered parts the author marked as \verdict{ANSWER\_PLUS\_CLAIM} and the judge did not. On re-reading, three were strengthened answers the haiku judge missed, three were close paraphrases the author over-flagged, in two the judge's reading was the better one (a dropped ``most'', and a covered part the model withheld), and two were borderline.
\end{itemize}
The pattern across rounds is consistent: telling ``answered'' from ``not answered'' is easy for both judge and person; telling ``answered'' from ``answered and slightly strengthened'' is hard for both. A single overall agreement figure hides this, and a figure measured under one rubric says nothing about the next.

\subsection{A standard faithfulness metric}
We scored the twenty hand-labelled responses from rounds 1 and 2 that were not pure refusals (October runs, judged under the October rubric) with RAGAS faithfulness \citep{es2024ragas} (ragas 0.4.3, claude-haiku-4-5 as its model), asking at response level whether the response contains at least one unsupported or added claim. Our judge agreed with the author on 19 of 20; RAGAS (score below 1) on 15 of 20. RAGAS scored every response below 1, including the five the author judged clean, and no threshold separates the groups. It counts as unsupported everything the context does not state, including advice, arithmetic on supplied figures, and ``I don't have that information''. This is not a flaw in RAGAS, which measures whether a response stays within its context; it does mean such metrics are unsuitable when the context is incomplete by design and withholding is the correct behaviour.

\subsection{Changing the judge}
Table~\ref{tab:judges} repeats all scoring with claude-sonnet-5-5 as judge, on the same responses. The stronger judge finds many more added claims in both models' answers, which suggests that most of the gap is general leniency in the haiku judge rather than self-preference. The gap is larger on haiku's own answers (0.25 to 0.56) than on llama's (0.26 to 0.41), which is consistent with some self-preference \citep{panickssery2024selfpreference}, but the two models write differently and this data cannot separate the two explanations. Most importantly, the comparison between the two models depends on the judge. On the default runs, with the haiku judge, claude-haiku-4-5 scores higher in \texttt{boundary\_accuracy} by 0.10 (paired bootstrap over the 30 cases, 95\% interval $[+0.01, +0.19]$). With the sonnet judge the difference is $-0.02$ $[-0.11, +0.08]$: the two models cannot be told apart. The first interval only just excludes zero, so this is a fragile result, but it shows that a model comparison resting on one judge may not survive a change of judge. Both judges are from the same model family; a judge from another family has not been tried.

\begin{table}[t]
\centering\small
\begin{tabular}{lcccccc}
\toprule
 & \multicolumn{2}{c}{\texttt{boundary\_accuracy}} & \multicolumn{2}{c}{\texttt{answer\_plus\_claim}} & \multicolumn{2}{c}{\texttt{unsupported\_claim}} \\
\cmidrule(lr){2-3}\cmidrule(lr){4-5}\cmidrule(lr){6-7}
Run & haiku & sonnet & haiku & sonnet & haiku & sonnet \\
\midrule
claude-haiku-4-5, default & 0.54 & 0.30 & 0.25 & 0.56 & 0.71 & 0.85 \\
llama3.1:8b, default & 0.44 & 0.32 & 0.26 & 0.41 & 0.90 & 0.93 \\
claude-haiku-4-5, grounded & 0.99 & 0.90 & 0.02 & 0.17 & 0.00 & 0.02 \\
llama3.1:8b, grounded & 0.87 & 0.81 & 0.08 & 0.19 & 0.06 & 0.09 \\
\bottomrule
\end{tabular}
\caption{The same responses scored by two judges. Columns headed haiku and sonnet name the judge model (claude-haiku-4-5, claude-sonnet-5-5). 95\% case-bootstrap intervals for every figure are in the repository (\texttt{experiments/rates\_9nov.json}).}
\label{tab:judges}
\end{table}

\section{Limitations}
This is a pilot. The dataset has 30 cases in three domains, written by one person. Only two models were tested, one of which is also the default judge. The hand labels come from one labeller, who also wrote the rubric; a second labeller has not yet been recruited. Both judges come from one model family. The grounded results for claude-haiku-4-5 are close to the ceiling, so in its current form the dataset cannot separate strong models that are given the instruction; harder cases, such as answers that must combine two pieces of the material, are needed for that. The knowledge packages describe rules that change over time; the evaluation treats them as fixed and does not measure real-world accuracy. The rubric changed several times during the work (a refusal verdict, then the added-claim verdict, then the boundary rules), and figures from different rubric versions are not comparable.

\section{Conclusion}
Scoring each part of a question separately shows two things a single score hides. When models are not told to stay within their material, they fill gaps freely; when they are told, the remaining failure is subtler: covered parts answered with the material made stronger than it is. This failure is also where LLM judges and people disagree most, and the choice of judge can decide whether one model looks better than another. We suggest that evaluations of this kind report per-category agreement with blind hand labels, repeat the hand check after every rubric or judge change, and report results under more than one judge.

\section*{Code and data}
Dataset, code, hand labels and per-run figures with intervals: \url{https://github.com/vaahiiid/partial-knowledge-eval} (MIT licence). The version used for this paper is tagged \texttt{paper-v1}.

\section*{Acknowledgements}
Kamran Biglari contributed tests and continuous integration and identified duplicated coverage in the dataset.

\bibliographystyle{plainnat}
\bibliography{refs}

@inproceedings{joren2025sufficient,
  title     = {Sufficient Context: A New Lens on Retrieval Augmented Generation Systems},
  author    = {Joren, Hailey and Zhang, Jianyi and Ferng, Chun-Sung and Juan, Da-Cheng and Taly, Ankur and Rashtchian, Cyrus},
  booktitle = {International Conference on Learning Representations (ICLR)},
  year      = {2025},
  note      = {arXiv:2411.06037}
}

@inproceedings{chen2024rgb,
  title     = {Benchmarking Large Language Models in Retrieval-Augmented Generation},
  author    = {Chen, Jiawei and Lin, Hongyu and Han, Xianpei and Sun, Le},
  booktitle = {Proceedings of the AAAI Conference on Artificial Intelligence},
  year      = {2024},
  note      = {arXiv:2309.01431}
}

@article{ming2024faitheval,
  title   = {{FaithEval}: Can Your Language Model Stay Faithful to Context, Even If ``The Moon is Made of Marshmallows''},
  author  = {Ming, Yifei and Purushwalkam, Senthil and Pandit, Shrey and Ke, Zixuan and Nguyen, Xuan-Phi and Xiong, Caiming and Joty, Shafiq},
  journal = {arXiv preprint arXiv:2410.03727},
  year    = {2024}
}

@article{zhang2026esb,
  title   = {Do {LLMs} Know When Evidence is Insufficient? {An} Evidence Sufficiency Benchmark for Answer-Abstention Calibration in Retrieval-Augmented Generation},
  author  = {Zhang, Hantian and Wu, Wentai},
  journal = {Computers, Materials \& Continua},
  volume  = {89},
  number  = {1},
  pages   = {69},
  year    = {2026},
  doi     = {10.32604/cmc.2026.086343}
}

@article{setiawan2026promptabstention,
  title   = {Prompt-Based Abstention Fails Under Misleading Context: A Controlled Study of Small Frozen {RAG} Models},
  author  = {Setiawan, Yohanes Andre},
  journal = {arXiv preprint arXiv:2608.22228},
  year    = {2026}
}

@article{wen2024knowyourlimits,
  title   = {Know Your Limits: A Survey of Abstention in Large Language Models},
  author  = {Wen, Bingbing and Yao, Jihan and Feng, Shangbin and Xu, Chenjun and Tsvetkov, Yulia and Howe, Bill and Wang, Lucy Lu},
  journal = {arXiv preprint arXiv:2407.18418},
  year    = {2024}
}

@article{kirichenko2025abstentionbench,
  title   = {{AbstentionBench}: Reasoning {LLMs} Fail on Unanswerable Questions},
  author  = {Kirichenko, Polina and Ibrahim, Mark and Chaudhuri, Kamalika and Bell, Samuel J.},
  journal = {arXiv preprint arXiv:2506.09038},
  year    = {2025}
}

@inproceedings{es2024ragas,
  title     = {{RAGAs}: Automated Evaluation of Retrieval Augmented Generation},
  author    = {Es, Shahul and James, Jithin and Espinosa Anke, Luis and Schockaert, Steven},
  booktitle = {Proceedings of the 18th Conference of the European Chapter of the Association for Computational Linguistics: System Demonstrations},
  pages     = {150--158},
  year      = {2024}
}

@inproceedings{niu2024ragtruth,
  title     = {{RAGTruth}: A Hallucination Corpus for Developing Trustworthy Retrieval-Augmented Language Models},
  author    = {Niu, Cheng and Wu, Yuanhao and Zhu, Juno and Xu, Siliang and Shum, Kashun and Zhong, Randy and Song, Juntong and Zhang, Tong},
  booktitle = {Proceedings of the 62nd Annual Meeting of the Association for Computational Linguistics (Volume 1: Long Papers)},
  pages     = {10862--10878},
  year      = {2024}
}

@inproceedings{panickssery2024selfpreference,
  title     = {{LLM} Evaluators Recognize and Favor Their Own Generations},
  author    = {Panickssery, Arjun and Bowman, Samuel R. and Feng, Shi},
  booktitle = {Advances in Neural Information Processing Systems},
  volume    = {37},
  year      = {2024}
}

@inproceedings{zheng2023judging,
  title     = {Judging {LLM}-as-a-Judge with {MT-Bench} and {Chatbot Arena}},
  author    = {Zheng, Lianmin and Chiang, Wei-Lin and Sheng, Ying and Zhuang, Siyuan and Wu, Zhanghao and Zhuang, Yonghao and Lin, Zi and Li, Zhuohan and Li, Dacheng and Xing, Eric P. and Zhang, Hao and Gonzalez, Joseph E. and Stoica, Ion},
  booktitle = {Advances in Neural Information Processing Systems 36, Datasets and Benchmarks Track},
  year      = {2023}
}

@article{norman2026reliability,
  title   = {Reliability without Validity: A Systematic, Large-Scale Evaluation of {LLM}-as-a-Judge Models Across Agreement, Consistency, and Bias},
  author  = {Norman, Justin D. and Rivera, Michael U. and Hughes, D. Alex},
  journal = {arXiv preprint arXiv:2606.19544},
  year    = {2026}
}

@misc{inspect2024,
  title        = {Inspect: A Framework for Large Language Model Evaluations},
  author       = {{UK AI Security Institute}},
  year         = {2024},
  howpublished = {\url{https://inspect.aisi.org.uk/}}
}
\end{document}